# A real-time decision support system for bridge management based on the rules generalized by CART decision tree and SMO algorithms[*]

### Shadi Abpeykar, Mehdi Ghatee[†]

Department of Computer Science, Amirkabir University of Technology, Tehran, Iran

**Abstract—** Under dynamic conditions on bridges, we need a real-time management. To this end, this paper presents a rule-based decision support system in which the necessary rules are extracted from simulation results made by Aimsun traffic micro-simulation software. Then, these rules are generalized by the aid of fuzzy rule generation algorithms. Then, they are trained by a set of supervised and the unsupervised learning algorithms to get an ability to make decision in real cases. As a pilot case study, Nasr Bridge in Tehran is simulated in Aimsun and WEKA data mining software is used to execute the learning algorithms. Based on this experiment, the accuracy of the supervised algorithms to generalize the rules is greater than 80%. In addition, CART decision tree and sequential minimal optimization (SMO) provides 100% accuracy for normal data and these algorithms are so reliable for crisis management on bridge. This means that, it is possible to use such machine learning methods to manage bridges in the real-time conditions.

1. Introduction

To manage the bridges, different decisions should be made; see e.g. [1] and [2] for fire prevention or [3] for incident management. For intelligent control of bridges, a decision support system (DSS) can be used in bridges. Such systems have been implemented for urban traffic management [4], hazmat transportation planning [5] and incident management [6]. Also for DSS in bridge management area, one can refer to [7], [8], [9], and [10].

---

[*] This paper is partially supported by Intelligent Transportation System Research Institute, Amirkabir University of Technology, Tehran, Iran.
[†] Corresponding author: Mehdi Ghatee, Associate Professor with Department of Computer Science, Amirkabir University of Technology, No 424, Hafez Ave, Tehran 15875-4413, Iran Fax: +98216497930, Email: ghatee@aut.ac.ir, URL: http://www.aut.ac.ir/ghatee

To make decisions based on historical databases, the supervised and the unsupervised learning algorithms can be used, see e.g., [5], [11], [12], [13] or [10]. In addition, a neural network for incident management was proposed by [14], [15], a decision tree for incident detection and its severity was developed by [16], [17] and some different algorithms for tunnel management were compared by [10]. However, decision making in bridges under dynamic situations, needs high accuracy and low process time. To this end, a neural network for damage detection in bridges was implemented by [18] and a SVM algorithm for automated bridges was developed by [19].

Following these works, in this paper a new bridge management DSS is proposed in which the extracted control rules from simulation software and historical data are combined. Then these rules are trained by suitable learning algorithms to generalize appropriate rules for the next real situations.

## 2. Bridge management decision support system

The framework of the bridge management DSS can be implemented by three subsystems:

- Simulation module
- If-then rule extraction module
- Learnable algorithms module

In simulation module, the different scenarios have been simulated in any micro-traffic simulators such as Aimsun, see e.g. [10]. Inputs and outputs of these simulations are considered in the second module to extract the necessary if-then rules. Then these rules are fed into WEKA software in two different types including "normal data" and "discrete data" to generalize the rules for the next situations, see similar work in [5]. Some supervised and unsupervised learning algorithms in the third module are applied to train the final set of if-then rules [20], [11], [21] or [22] to see some advanced versions of these algorithms. In this module, to find the most accurate results in a reasonable process time, different algorithms are compared. The details of the framework of this DSS are presented in Fig. 1 . As one can see, based on the real-time traffic data, some scenarios are made in Aimsun simulator, then some if-then rules are made by the aid of fuzzy rule generation algorithms and then these rules are fed to learning algorithms to find the most accurate and the fastest solutions. The rules for Advanced Traffic Management System (ATMS) and Advanced Traveler Information System (ATIS) are saved to control the bridge in real-time. The results are also saved in the bridge documentation subsystem as historical data for the next experiments.

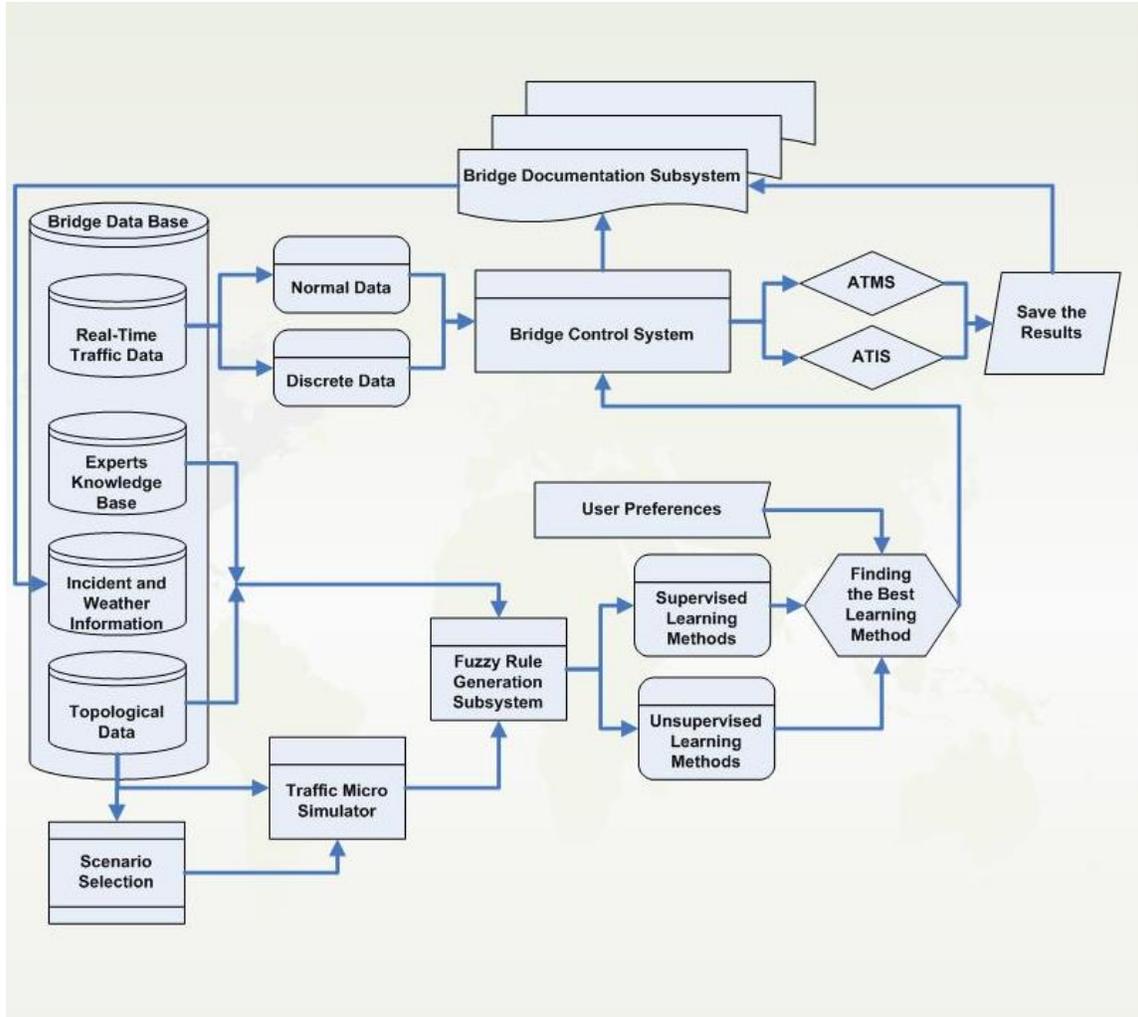

*Fig. 1    Bridge management DSS framework*

3. Case study and simulation results

Nasr Bridge is one of the most important bridges in Tehran depicted in Fig. 2. This bridge located between two other bridges and so controlling the traffic jam through this bridge is very important. To control this bridge, a great number of scenarios in Aimsun simulation environment are defined with respect to the different weather conditions (including snowing, raining, fogy and wet), various traffic flows (weekday and holiday travels), three time intervals (one hour at the morning, one hour at the noon and one hour at the night) and four seasons.

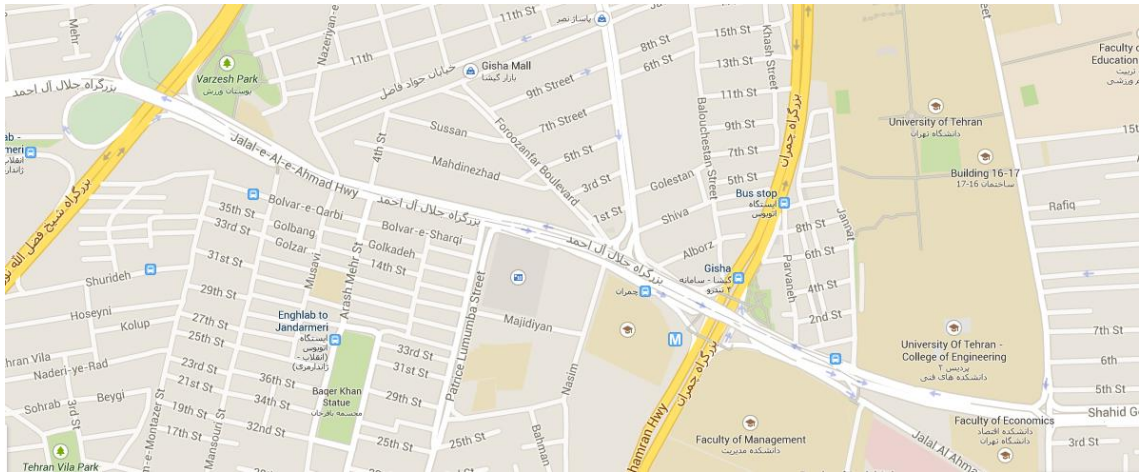

*Fig. 2      Nasr bridge for simulation experiments (https://www.google.com/)*

For these scenarios, some incident situations are simulated and then the traffic parameters are saved. To evaluate the effect of advanced traffic management system (ATMS) and advanced traveler information system (ATIS) on Nasr bridge, without losing from generality and just for simplicity, we assume the following two modules for simulation:

- M1: Ramp metering
- M2: Rerouting

These modules are simulated in Aimsun version 7 simulation environment and the outputs of the simulator are used to train the effects of ATMS and ATIS decisions by learning machine algorithms. The objective function to choose the best decisions are reducing travel time and delay time for all of the drivers. After simulation, the input data and the output of simulation are fed to fuzzy if-then rule generator algorithm and then the generated if-then rules are fed to WEKA[‡] data-mining software for rule generalization. In WEKA, there are two types of filters: normal and discrete. Both of these filters are used in this experiment to test the improvement rate in the quality of learning machine algorithms.

‡http://wiki.pentaho.com/

In Nasr bridge case study, 11520 new if-then rules are extracted from WEKA and they are used to train the supervised and the unsupervised learning algorithms in order to generalize the rules for real situations.

In supervised learning algorithms, the following algorithms are considered:

- Feed Forward neural network [23].
- C4.5 decision tree [24].
- Naive Bayes Tree (NB Tree) [24].
- Classification and Regression Tree (CART) [25], [26].
- Naive Bayes [27].
- Sequential Minimal Optimization (SMO) [27].
- Hidden Naive Bayes (HNB) [28].
- Support Vector Machine (SVM) [27], [29].

In addition, the following unsupervised learning algorithms are considered:

- Expectation-Maximization (EM) [30].
- K-Mean clustering algorithm [31].
- Farthest First [32].
- Learning Vector Quantization (LVQ) [33].
- Hierarchical Clustering [34].
- Filtered Clustering [34].
- Self-Organization Map (SMO) [35], [36].

To evaluate the accuracy of these learning algorithms, the extracted if-then rules are partitioned into two sets. The first set is used to train and the other set is used to test and to evaluate. The rates of division of the rules are as follows:

- Training subset: 65% and Test subset: 35%
- Training subset: 70% and Test subset: 30%
- Training subset: 80% and Test subset: 20%

Based on the results of the unsupervised algorithms in Fig. 3, the most accuracy corresponds to the self-organization map (SOM) with 68.7%. In addition, the results of discrete filter are better than normal filter. Also by comparing the process time of these algorithms, it is shown that the farthest first algorithm is better than other unsupervised algorithms for normal and discrete filters. See Fig. 4 for normal filter results.

In Fig. 5 the performance of the supervised algorithms are compared. The illustrated algorithms generalize the rules with the accuracy greater than 80% for normal and discrete data, which is very interesting result. This means that all of these algorithms can be used for decision making with a high accuracy for bridge management. Although between these algorithms CART decision tree and sequential minimal optimization (SMO) provides 100% accuracy for normal data and so these algorithms are so reliable for crisis management on bridge. On the other hand, hidden Naive Bayes obtains the most accuracy of the supervised learning algorithms on discrete data.

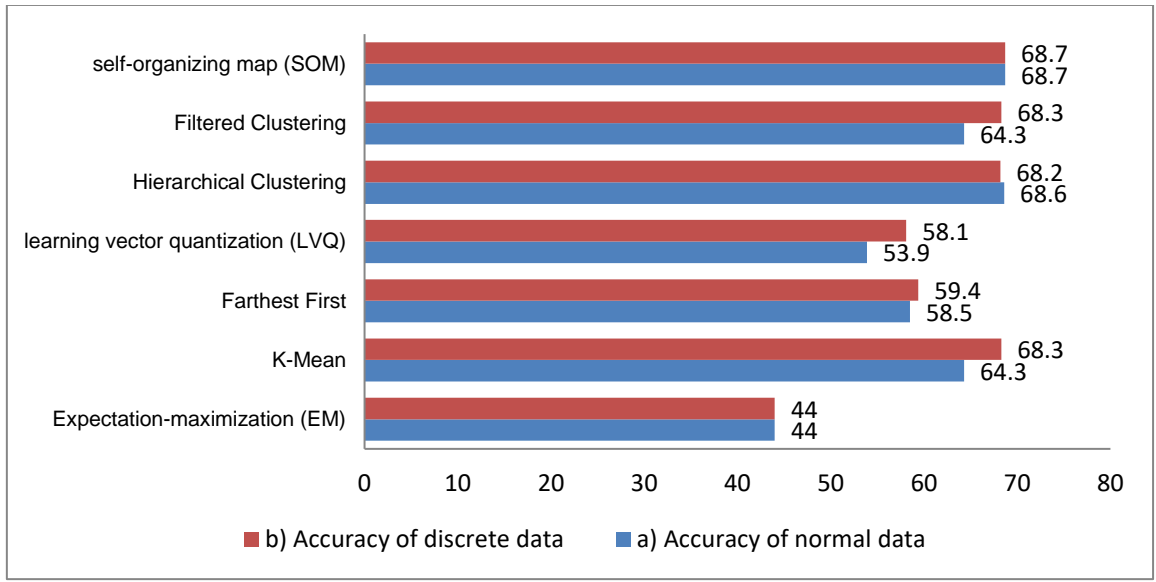

*Fig. 3        Comparison between accuracy of unsupervised learning algorithms on data of Nasr bridge*

By comparing the supervised learning algorithms based on the process time, it is demonstrated that the process time of all of these algorithms is less than 1 second, which shows that they can be used for real-time implementation.

++++++++++++++++++++++++++++++++++++++++++

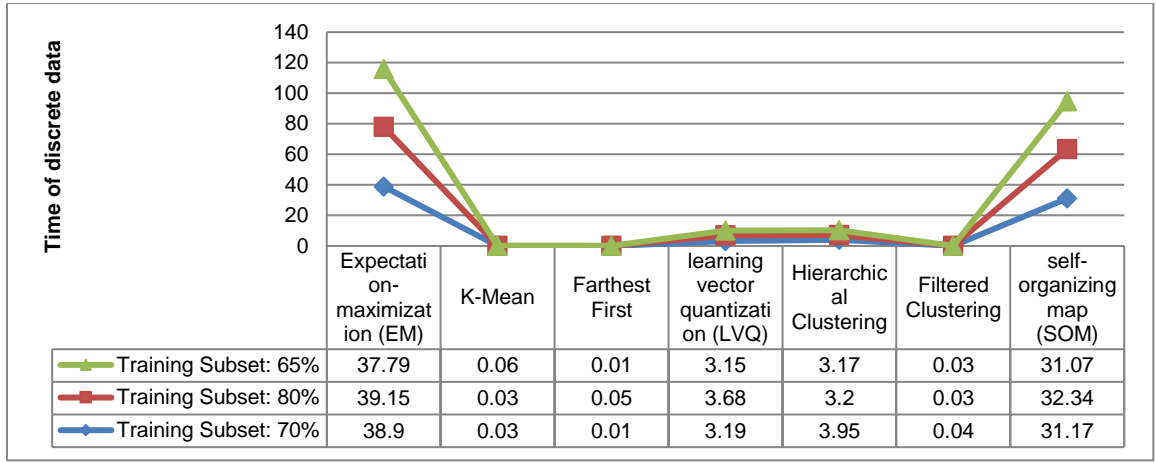

| | Expectation-maximization (EM) | K-Mean | Farthest First | learning vector quantization (LVQ) | Hierarchical Clustering | Filtered Clustering | self-organizing map (SOM) |
|---|---|---|---|---|---|---|---|
| Training Subset: 65% | 37.79 | 0.06 | 0.01 | 3.15 | 3.17 | 0.03 | 31.07 |
| Training Subset: 80% | 39.15 | 0.03 | 0.05 | 3.68 | 3.2 | 0.03 | 32.34 |
| Training Subset: 70% | 38.9 | 0.03 | 0.01 | 3.19 | 3.95 | 0.04 | 31.17 |

*Fig. 4    Comparison between training time of unsupervised learning algorithms on data of Nasr bridge with discrete filter*

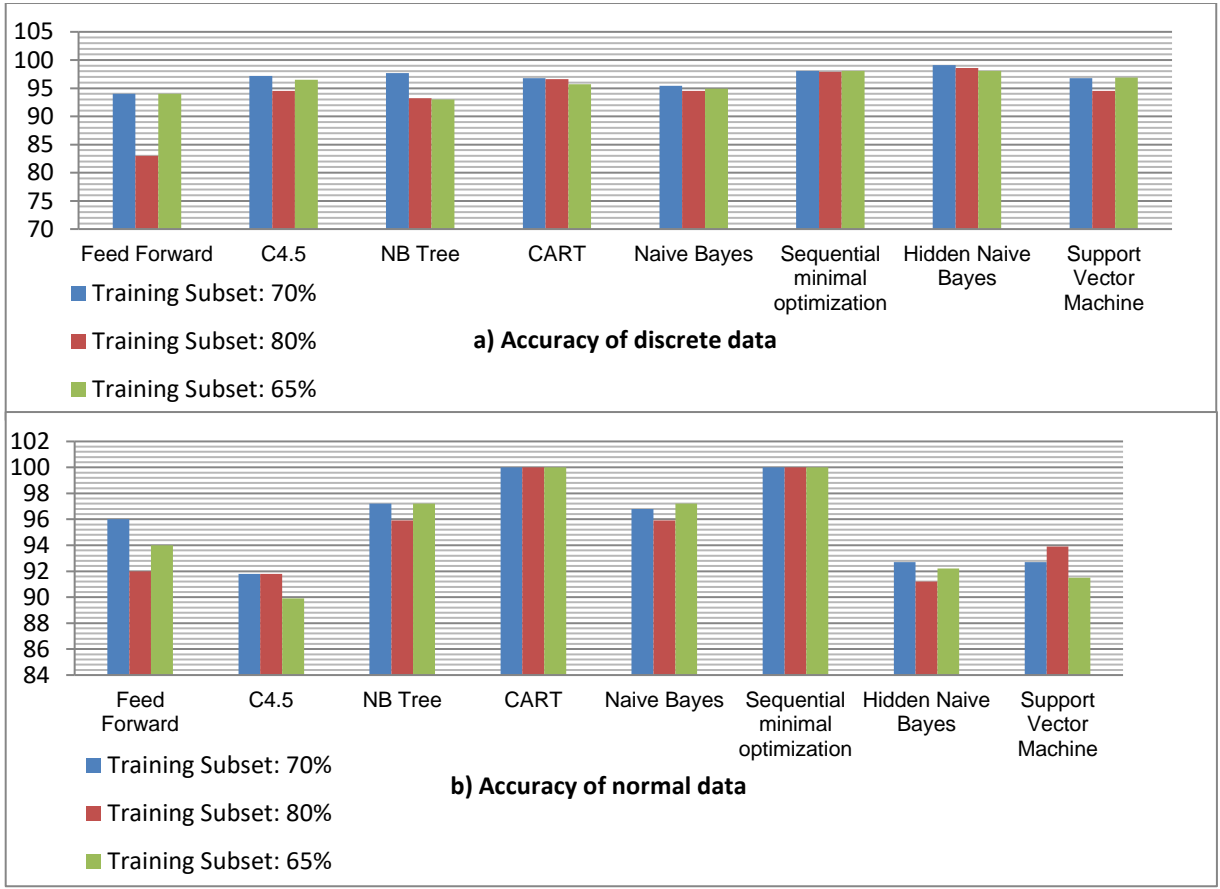

*Fig. 5    Comparison between accuracy of supervised learning algorithms on*
*data of Nasr bridge*
*a)  Discrete data ,*
*b)  Normal data*

4.  Conclusions

A new decision support system for bridge management has been proposed in this paper. Firstly, the bridge is simulated in a traffic micro-simulator such as Aimsun, then the simulation results are stated as some if-then rules by fuzzy if-then rule generator algorithm and finally the rules are trained and generalized by CART decision tree or sequential minimal optimization (SMO) algorithms. Then in real situations, it is possible to manage the bridge based on the generalized rules. We showed the applicability of this approach in Nasr bridge of Tehran. We found 100% accuracy in less than 1 second processing time in our case study. This approach can be customized for controlling the different transportation infrastructures.